\documentclass[10pt, conference, compsocconf]{IEEEtran}

\fontfamily{cmr}\selectfont
\usepackage[utf8]{inputenc}

\ifCLASSOPTIONcompsoc
  \usepackage[caption=false,font=normalsize,labelfont=sf,textfont=sf]{subfig}
\else
  \usepackage[caption=false,font=footnotesize]{subfig}
\fi

\usepackage{xcolor}
\usepackage{amsmath}
\usepackage{amssymb}
\usepackage{mathtools}
\usepackage{graphicx}
\usepackage{cite}

\renewcommand\IEEEkeywordsname{Keywords -}

\begin{document}

\title{HoughNet: neural network architecture for vanishing points detection}



\author{\IEEEauthorblockN{Alexander Sheshkus\IEEEauthorrefmark{1} \IEEEauthorrefmark{2} \IEEEauthorrefmark{3}
Anastasia Ingacheva\IEEEauthorrefmark{2} \IEEEauthorrefmark{6},
Vladimir Arlazarov \IEEEauthorrefmark{2} \IEEEauthorrefmark{4} and
Dmitry Nikolaev \IEEEauthorrefmark{2} \IEEEauthorrefmark{5}}

\IEEEauthorblockA{\IEEEauthorrefmark{1}Email: astdcall@gmail.com, Phone: +7 (905) 569-85-08}
\IEEEauthorblockA{\IEEEauthorrefmark{2}LLC Smart Engines Service, 117312, Moscow, Russia.}
\IEEEauthorblockA{\IEEEauthorrefmark{3}Federal Research Center ”Computer Science and Control” RAS, 119333, Moscow, Russia.}
\IEEEauthorblockA{\IEEEauthorrefmark{4}Moscow Institute of Physics and Technology (National Research University), 117303, Moscow, Russia.}
\IEEEauthorblockA{\IEEEauthorrefmark{5}Institute for Information Transmission Problems (Kharkevich Institute) RAS, 127051, Moscow, Russia}
\IEEEauthorblockA{\IEEEauthorrefmark{6}National Research University Higher School of Economics, 101000, Moscow, Russia.}
}

\maketitle

\begin{abstract}
In this paper we introduce a novel neural network architecture based on Fast Hough Transform layer. The layer of this type allows our neural network to accumulate features from linear areas across the entire image instead of local areas. We demonstrate its potential by solving the problem of vanishing points detection in the images of documents. Such problem occurs when dealing with camera shots of the documents in uncontrolled conditions. In this case, the document image can suffer several specific distortions including projective transform. To train our model, we use MIDV-500 dataset and provide testing results. Strong generalization ability of the suggested method is proven with its applying to a completely different ICDAR 2011 dewarping contest. In previously published papers considering this dataset authors measured quality of vanishing point detection by counting correctly recognized words with open OCR engine Tesseract. To compare with them, we reproduce this experiment and show that our method outperforms the state-of-the-art result.
\end{abstract}


%
\IEEEpeerreviewmaketitle

\section{Introduction}
\label{intro}
\begin{figure}[bt]
\centering
\includegraphics[height=4.5cm]{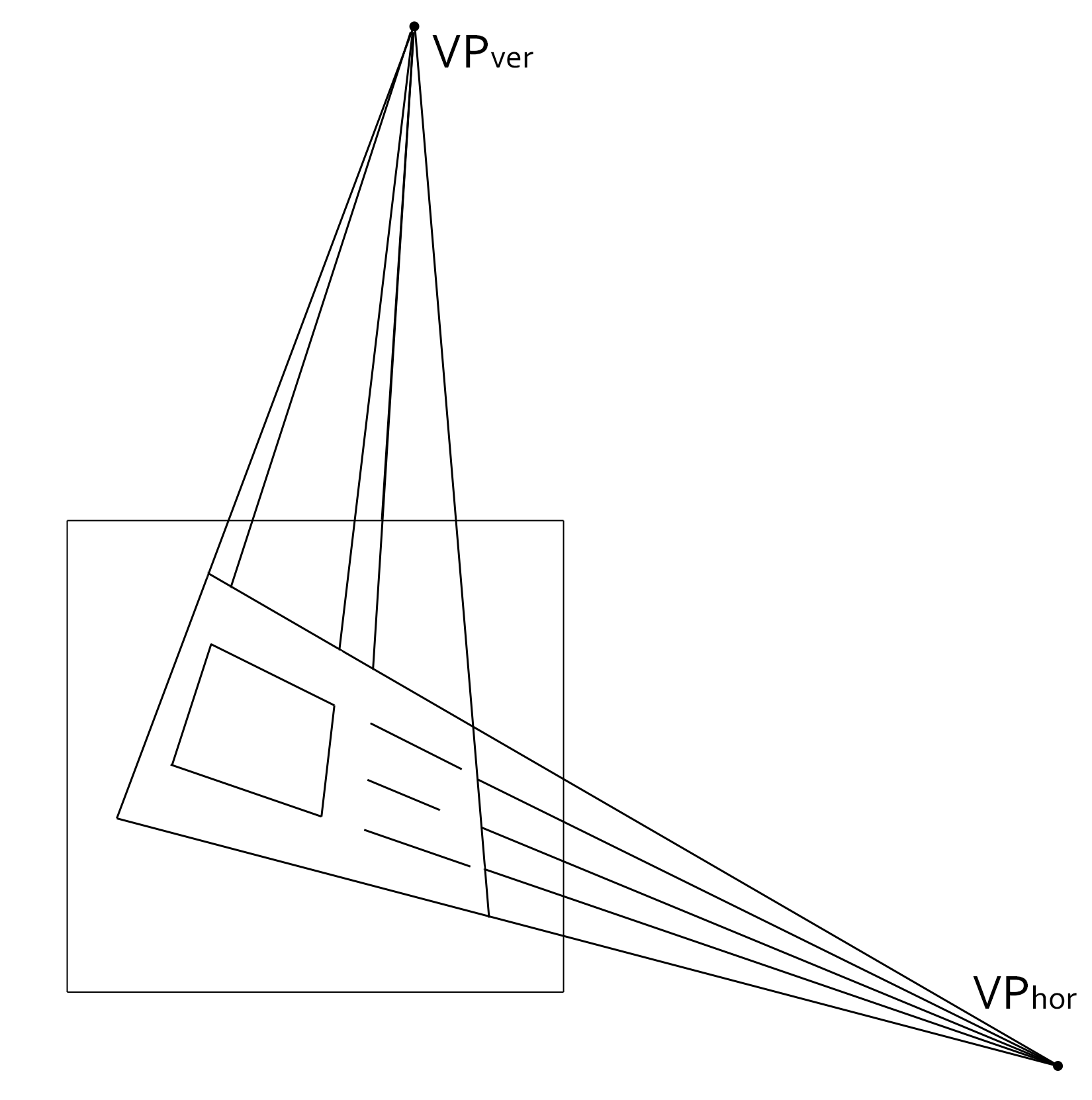}
\caption{Horizontal and vertical vanishing points for the document.}
\label{VP_example}
\end{figure}

Mobile photo technologies development and ubiquitous mobile devices capable of making images of acceptable for text recognition quality led to the fact that various documents are being photographed for digital processing. Since these images are taken in uncontrolled conditions they suffer from various camera-specific distortions including the projective ones. Therefore, we have to solve the task of the document rectification prior to any other analysis. The purpose of this step is to find an image transformation which will make the document rectangular and correctly oriented. Even though recognition algorithms which work for distorted images \mbox{exist}~\cite{end-to-end}, the mainstream methods include this step since it makes most of the following document understanding simpler. To rectify the document, we have to calculate the parameters of the homography matrix. It is possible if we know the correspondence between four points (two coordinates for each) in the original and rectified image. Selection of these points is possible in different ways. One of the choices is to find the document quadrangle itself using one of the many known methods \mbox{\cite{doc-loc, borders1, borders2}} and then transform it to a rectangular shape. This approach is not always applicable since the borders of the document may be obscured with irrelevant objects, mixed with a background or even located outside of the frame. The alternative is to find two vanishing points. This method works for documents since they typically contain strings of text in a specific direction and therefore it is possible to evaluate horizontal and vertical vanishing points (see Fig. \ref{VP_example}). Using these vanishing points one can calculate the homography matrix and rectify the image except for image shift and scale. 

Vanishing points detection is a commonly known problem which was addressed multiple times in multiple contexts. A classical approach to vanishing points detection is demonstrated in Barnards work \cite{barnard-old}. He uses Gaussian sphere centered in the origin of the coordinates while image plane is placed at some distance from image plane -- focal length ($z$ coordinate of the image plane). Every point on the image can be mapped to a point on the sphere and then be treated as a radius-vector. With this trick, one can map points at infinity to the finite space and deal with them using regular methods. To detect vanishing points we have to find the intersection points of all lines in the image and then merge them into clusters. All lines which belong to the same cluster represent a bunch of parallel lines in some perspective. Such technique is used not only for vanishing points detection: for example, in \cite{text-sphere} authors suggest to find text baselines and skewness of the symbols using clusters of points on the Gaussian sphere. But still, these algorithms have task-specific parameters and therefore lack general robustness. In \cite{gnomonic} Gaussian sphere is also used, but as an input for a deep convolutional neural network for vanishing point detection. The main advantage of this method is that it is possible to generate synthetic training data since neural network uses projections on the sphere instead of input images. The problem of this approach is that line segments in the input image still have to be found somehow.

With algorithms evolution and specialization three essentially different cases for the problem emerge. Vanishing point detection method in a road scene (while analyzing images taken with video registrator) typically has to find the target point within the image \cite{vp-on-road} while in the ``Manhattan world''~\cite{manh-world} three orthogonal vanishing points exist. When dealing with documents, we expect to find two vanishing points which are located outside of the image. Generally, it is possible to capture the document otherwise, but we will not consider this case since distortions are very strong and the contained data is hardly recognizable even after the rectification procedure. Since we are looking for vanishing points outside the image it is impossible to use direct convolutional neural network approach which becomes popular in road scene vanishing point detection \cite{kor-road}. 

Despite the fact that there are methods for vanishing point detection using intersection points clusterization \cite{line-clustering}, the base method for the task relies on double Hough Transform \cite{hough-clear}. This method is simple and clear but too non-robust to different image distortions and corruptions and is applicable only to high-quality input data. Most of the algorithms for vanishing point detection more or less are based on this approach. Since Hough Transform is an integral operator on the image, it is worth mentioning that integral operators are used in a wide range of algorithms from skew angle calculation \cite{text-block} to images reconstruction in computed tomography \cite{tomo-int}. 

In paper \cite{radon-1} authors use direct and inverse Radon transformations combination to calculate candidates for vanishing points and there is an improvement to this method in \cite{radon-2} with a usage of RANSAC scheme. The results are promising, but we can see, that the method is highly dependent on the text amount and, more importantly, that the vertical vanishing point is less confident since there are much fewer vertical lines in the document. 

In \cite{myhough-icmv16} authors train a recognition neural network using the result of Hough Transform \cite{ht} as a feature map. In \cite{koppen} authors describe the method for Hough transform calculation with neural network and in \cite{milletari} authors use Hough voting procedure to use not only the NN answers, but also descriptors from the second-last fully connected layer. In paper \cite{myhough-icmv17} there is a definition of a Fast Hough Transform (FHT) layer as a linear operator for feature space transformation. We will develop this idea further and introduce new neural network architecture based on this type of layer. 

\section{Vanishing Points detection using FHT}
\label{thetask}
In this paper, we consider the problem of vanishing point detection in the images of documents using FHT which has a lot of applications \cite{fht}. These images typically contain two vanishing points -- the first one is obtained from the top and bottom borders along with text strings, the second one emerges from the left and right borders of the document along with the borders of some elements in the document content. Specifically, we will consider only the cases when both vanishing points are located outside the image.




FHT algorithm calculates four parts of the output image separately. These parts corresponds to angle ranges $[-45^\circ, 0^\circ]$, $[0^\circ, 45^\circ]$, $[45^\circ, 90^\circ]$ and $[90^\circ, 135^\circ]$ respectively if the input image is a square. We will refer to them as $H_1$, $H_2$, $H_3$ and $H_4$. We also will use $H_{12}$ and $H_{34}$ for vertically joined results of $H_1$ with $H_2$ and $H_3$ with $H_4$ respectively. Consequent FHT appliances will be referred to $H_m(H_n)$ where $n$ and $m$ specify the angle range for the first and the second transformations respectively.

\begin{figure}[bt]
\centering
\subfloat[Input Image]{\includegraphics[height=2.5cm]{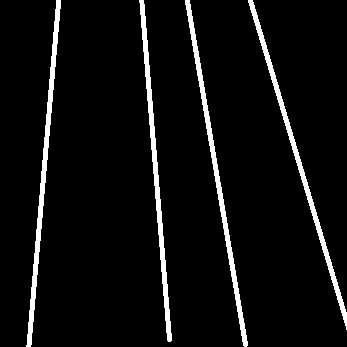}\label{fht-ex1_a}}
\hspace{0.5em}
\subfloat[$H_{12}$]{\includegraphics[height=2.5cm]{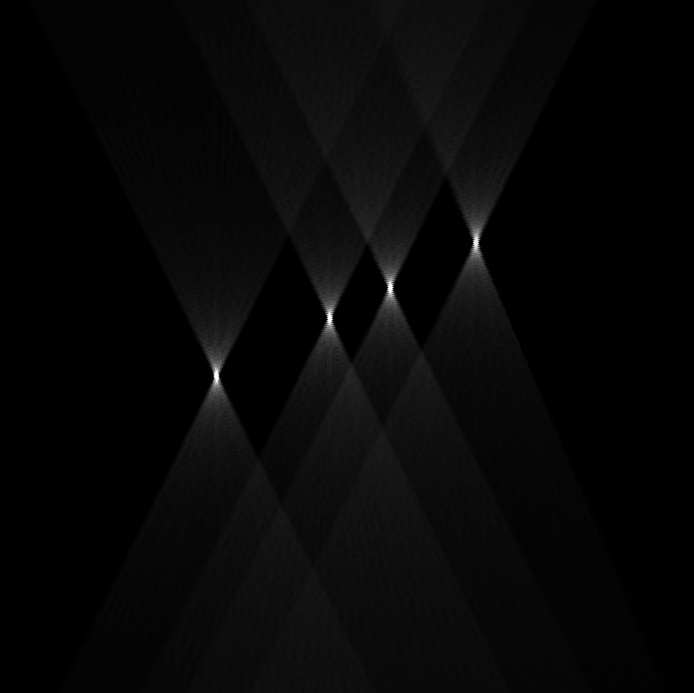}\label{fht-ex1_b}}
\hspace{0.5em}
\subfloat[$H_{34}(H_{12})$]{\includegraphics[height=2.5cm]{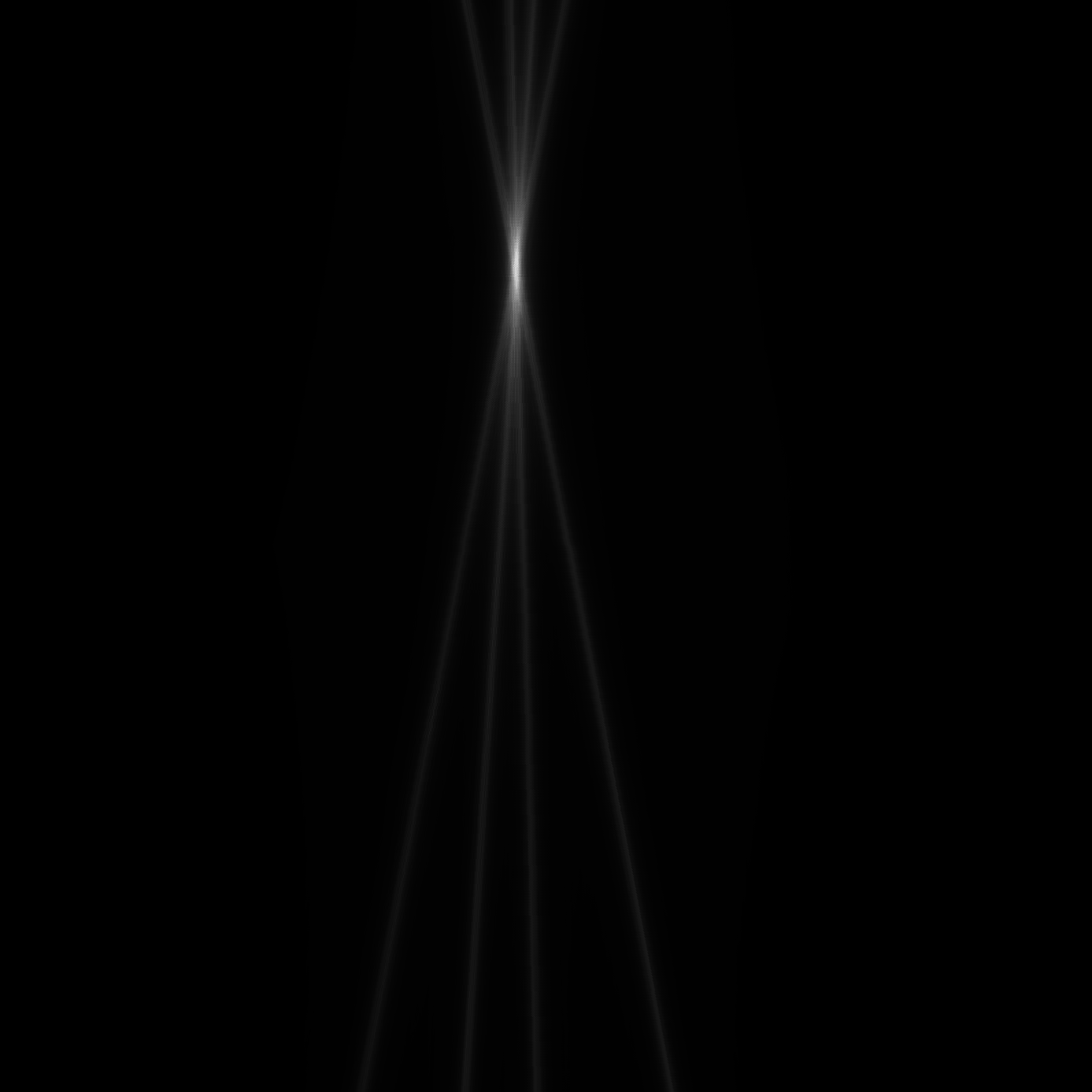}\label{fht-ex1_c}}
\caption{Hough transform for Vanishing Point detection.}
\label{fht-ex1}
\end{figure}

Every line in the input image transforms to a specific point in the first FHT image. For the lines intersecting in one point the corresponding points in the first FHT image will be collinear. The line containing all these points will transform to a point in the consequent FHT image. For a better understanding of the basics of the suggested method, we consider an example. In Fig. \ref{fht-ex1_a} there are four lines which all intersect in one point somewhere far above the image. If we calculate $H_{12}$ from this image we obtain \ref{fht-ex1_b}, where every local maximum corresponds to a certain line in the input image. Now we calculate $H_{34}$ from the image with points and obtain Fig. \ref{fht-ex1_c}. There is only one local maximum and it represents the intersection point of lines on the input image. In our problem we deal only with the vanishing points outside the image, otherwise we would consider $H_{12}$ for the second Hough operator.

\section{Suggested approach}
\label{sapproach}

Our solution has to work correctly with vanishing points outside the image including the points at infinity. To overcome difficulties of this case we suggest a new neural network architecture based on the FHT layer introduced in \cite{myhough-icmv17}. After the vanishing points detection we perform the rectification procedure as described in \cite{radon-1}. 

\subsection{Fast Hough Transform layer}
\label{fht-layer}
Our neural network architecture is based on the FHT layer which performs transformation of the specific angle range. Since this layer is a linear operator, back propagating gradient through it does not require any additional effort. We also want to mention that this layer has no trainable coefficients and is needed only for feature maps transformation. For $H_{12}$ (which corresponds to mostly vertical lines on the input image) we can calculate a point $(s, \alpha)$ from line segment $(x_0, 0) - (x_1, h)$ using equations:

\begin{equation}
\label{l2p-hor}
\left\{ 
\begin{array}{l}
s=x_0+d_v(\alpha) \\
\alpha=h-(x_1-x_0)
\end{array}
\right.
\end{equation}

For $H_{34}$ line segment is defined as $(0, y_0) - (w, y_1)$ and $(s, \alpha)$ can be calculated with:

\begin{equation}
\label{l2p-ver}
\left\{ 
\begin{array}{l}
s=y_0+d_h(\alpha) \\
\alpha=w-(y_0-y_1)
\end{array}
\right.
\end{equation}

In equations (\ref{l2p-hor}), (\ref{l2p-ver}) $d_v(\alpha)$ and $d_h(\alpha)$ represent image skewing \cite{deskew-fht} and can be written as:

\begin{equation}
\label{skew}
\left\{ 
\begin{array}{l}
d_v(\alpha)=h-\alpha/2 \\
d_h(\alpha)=\alpha/2
\end{array}
\right.
\end{equation}

\begin{figure}[bt]
\centering
\subfloat[Input image]{\includegraphics[height=2.5cm]{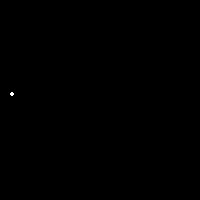}\label{skew-deskew_a}}
\hspace{0.5em}
\subfloat[Without skew]{\includegraphics[height=2.5cm]{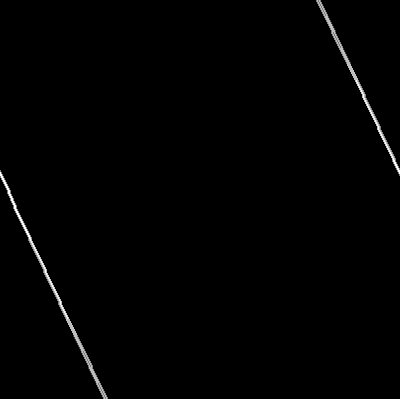}\label{skew-deskew_b}}
\hspace{0.5em}
\subfloat[With skew]{\includegraphics[height=2.5cm]{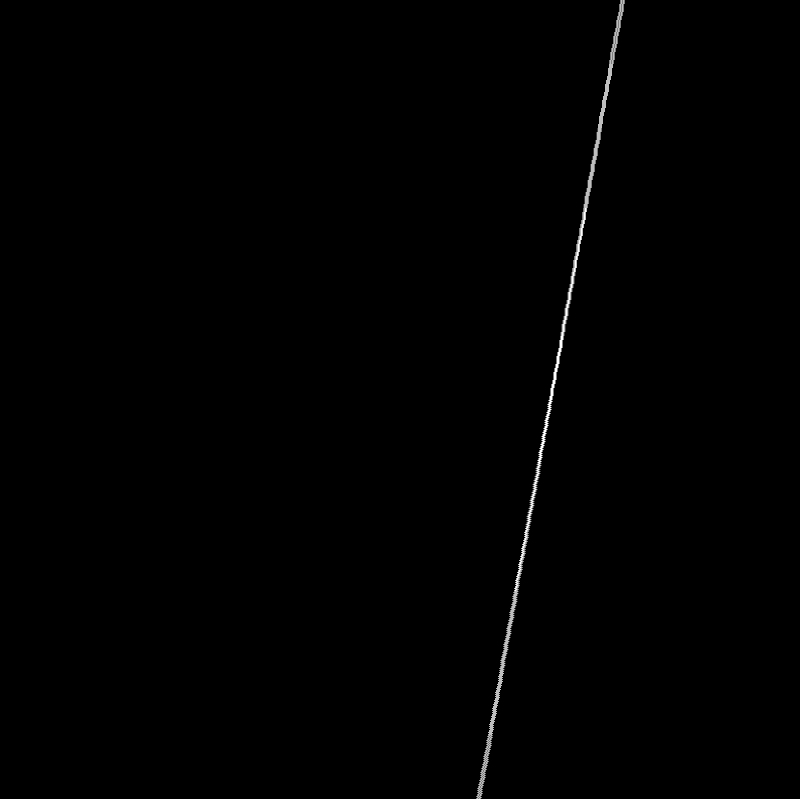}\label{skew-deskew_c}}
\caption{Examples of $H_{12}$ with zero skew and with skew by eq. (\ref{skew}).}
\label{skew-deskew}
\end{figure}

This skewing is useful since it makes the feature map continuous. In Fig. \ref{skew-deskew} there is an example of a simple input image \ref{skew-deskew_a}, $H_{12}$ image with zero skew \ref{skew-deskew_b} and with skew according to equations (\ref{skew}) \ref{skew-deskew_c}.

One can evaluate the correspondence of points and lines between the input and the resulting feature maps. Every point in the input image transforms into different lines on $H_{12}$ and $H_{34}$ parts of the output image. If we have a point $(x,y)$ in the input image, then the corresponding line on the Hough image can be calculated according to the equations:

\begin{equation}
\label{p2l-hor}
\left\{ 
\begin{array}{l}
s = x + \cfrac{(\alpha - h)y}{h} + d_v(\alpha) \\
\alpha=[0; 2h-1]
\end{array}
\right.
\end{equation}

\begin{equation}
\label{p2l-ver}
\left\{ 
\begin{array}{l}
s = y - \cfrac{(\alpha - w)x}{w} + d_h(\alpha) \\
\alpha=[0; 2w-1]
\end{array}
\right.
\end{equation}

\subsection{Vanishing points evaluation}
\label{vp-ev}

Using equations (\ref{l2p-hor}), (\ref{l2p-ver}), (\ref{skew}), (\ref{p2l-hor}), (\ref{p2l-ver}) it is possible to calculate the correspondence between points in coordinate space of $I$ and in coordinate space of $H_m(H_n)$:

\begin{equation}
\label{p2l-ver-l2p}
\left\{ 
\begin{array}{l}
s'=\cfrac{3h}{2} + \cfrac{4hx-w(2y+h)}{2(h-2y)} \\
\alpha'=\cfrac{(2y+h)(w+h)}{2y-h}
\end{array}
\right.
\end{equation}

\begin{equation}
\label{p2l-hor-l2p}
\left\{ 
\begin{array}{l}
s'=\cfrac{3w}{2} + \cfrac{4wy+h(2x-3w)}{2(2x-w)} \\
\alpha'=\cfrac{(2x-3w)(w+h)}{2x-w}
\end{array}
\right.
\end{equation}

Equation (\ref{p2l-ver-l2p}) is for $H_{34}(H_{12})$ while equation (\ref{p2l-hor-l2p}) is for $H_{34}(H_{34})$.

With known $s'$ and $\alpha'$ one can evaluate points coordinates in the original image coordinates system with equations (for $H_{34}(H_{12})$ and $H_{34}(H_{34})$):
 
\begin{equation}
\label{back-ver}
\left\{ 
\begin{array}{l}
x=\cfrac{\alpha' w +(3h-2s')(w+h)}{2(\alpha'-(w+h))} \\
y=\cfrac{h(\alpha'+w+h)}{2(\alpha' - (w+h))}
\end{array}
\right.
\end{equation}

\begin{equation}
\label{back-hor}
\left\{ 
\begin{array}{l}
x=\cfrac{w(\alpha' - 3(w+h))}{2(\alpha'-(w+h))} \\
y=\cfrac{\alpha' h +(3w-2s')(w+h)}{2(\alpha'-(w+h))}
\end{array}
\right.
\end{equation}


\section{HoughNet NN architecture}
\label{nnarch}
The main purpose of the proposed neural network architecture is to transform feature space and allow convolutional layers to operate with linear areas across the image instead of local areas. When considering the vanishing point detection problem, motivation for such transformation is that generally, one cannot solve the task operating only with local areas. On the contrary, this task can be solved if we operate with linear objects and their properties. Hence, neural network based on the FHT layer seems perfectly reasonable for the task. 

For the Hough transform it is proved, that there is no single transformation for the entire range of angles \cite{2bexplain}, therefore we suggest to build a two-branched neural network (for vertical and horizontal vanishing points). Every branch consists of three convolutional blocks and two FHT layers between them. \mbox{Table}~\ref{table:houghnet-arch} contains detailed layers description for both branches which are identical except for the fourth layer. The total number of trainable parameters is $31196$ per branch.

Our neural network architecture is mostly inspired with these points: \begin{itemize}
    \item convolution layers between Hough layers represent peak detectors;
    \item line detection with Hough transformation in presence of the noise and outliers in the data implies usage of convolutions \cite{kiryati};
    \item multichannel Hough map followed with non linear function approximator allows NN not only operate with accumulated value along the given line but also with its statistics, for example dispersion.
\end{itemize}

\begin{table}[!t]
	\caption{HoughNet architecture}
    \label{table:houghnet-arch}
	\begin{tabular}{|p{0.015\textwidth}|p{0.03\textwidth}|p{0.25\textwidth}|p{0.07\textwidth}|} 
	\hline
	\multicolumn{4}{|c|}{\textbf{Layers}} \\
	\hline
		\# & Type & Parameters & Activation function\\ 
		\hline
		1 &  conv & 12 filters 5$\times$5, stride 1$\times$1, no padding & $relu$\\ 
		\hline
		2 &  conv & 12 filters 5$\times$5, stride 2$\times$2, no padding & $relu$\\ 
		\hline
		3 &  conv & 12 filters 5$\times$5, stride 1$\times$1, no padding & $relu$\\ 
		\hline
		4 &  FHT & $H_{12}$ for vertical, $H_{34}$ for horizontal & -\\ 
		\hline
		5 &  conv & 12 filters 3$\times$9, stride 1$\times$1, no padding & $relu$\\ 
		\hline
		6 &  conv & 12 filters 3$\times$5, stride 1$\times$1, no padding & $relu$\\ 
		\hline
		7 &  conv & 12 filters 3$\times$9, stride 1$\times$1, no padding & $relu$\\ 
		\hline
		8 &  conv & 12 filters 3$\times$5, stride 1$\times$1, no padding & $relu$\\ 
		\hline
		9 &  FHT & $H_{34}$ for both branches & -\\ 
		\hline
		10 &  conv & 16 filters 5$\times$5, stride 3$\times$3, no padding & $relu$\\ 
		\hline
		11 &  conv & 16 filters 5$\times$5, stride 3$\times$3, no padding & $relu$\\ 
		\hline
		12 &  conv & 1 filter 5$\times$5, stride 1$\times$1, no padding & $1-rbf$\\ 
		\hline
	\end{tabular}
\end{table}

The neural network yields two images, one for the vertical vanishing point and one for the horizontal vanishing point. From every image we take a point with maximum intensity as an answer and transform its coordinates back to the original image coordinates space using equations (\ref{back-ver}) and (\ref{back-hor}) mixed with coordinates transform according to convolution layers.

The neural network is trained with minimization of $L2$ distance between the given and an ideal answer. For the ideal answer we used zero-filled images with the one-filled rectangle of $5\times5$ at the position of the correct answer. Convergence rate is very low and we were training our neural network for $15000$ epochs. Even though the amount of epochs is drastic, the training process took about $10$ days on a single-GPU PC which is acceptable. Moreover, this neural network can be quite universal and will not require retraining from scratch for every new case. In our future works, we plan to deal with that problem and develop a new cost function which would reduce the number of required epochs.

\section{Used Datasets}
\label{data}

\begin{figure}[bt]
\centering
\subfloat{\includegraphics[width=0.15\textwidth]{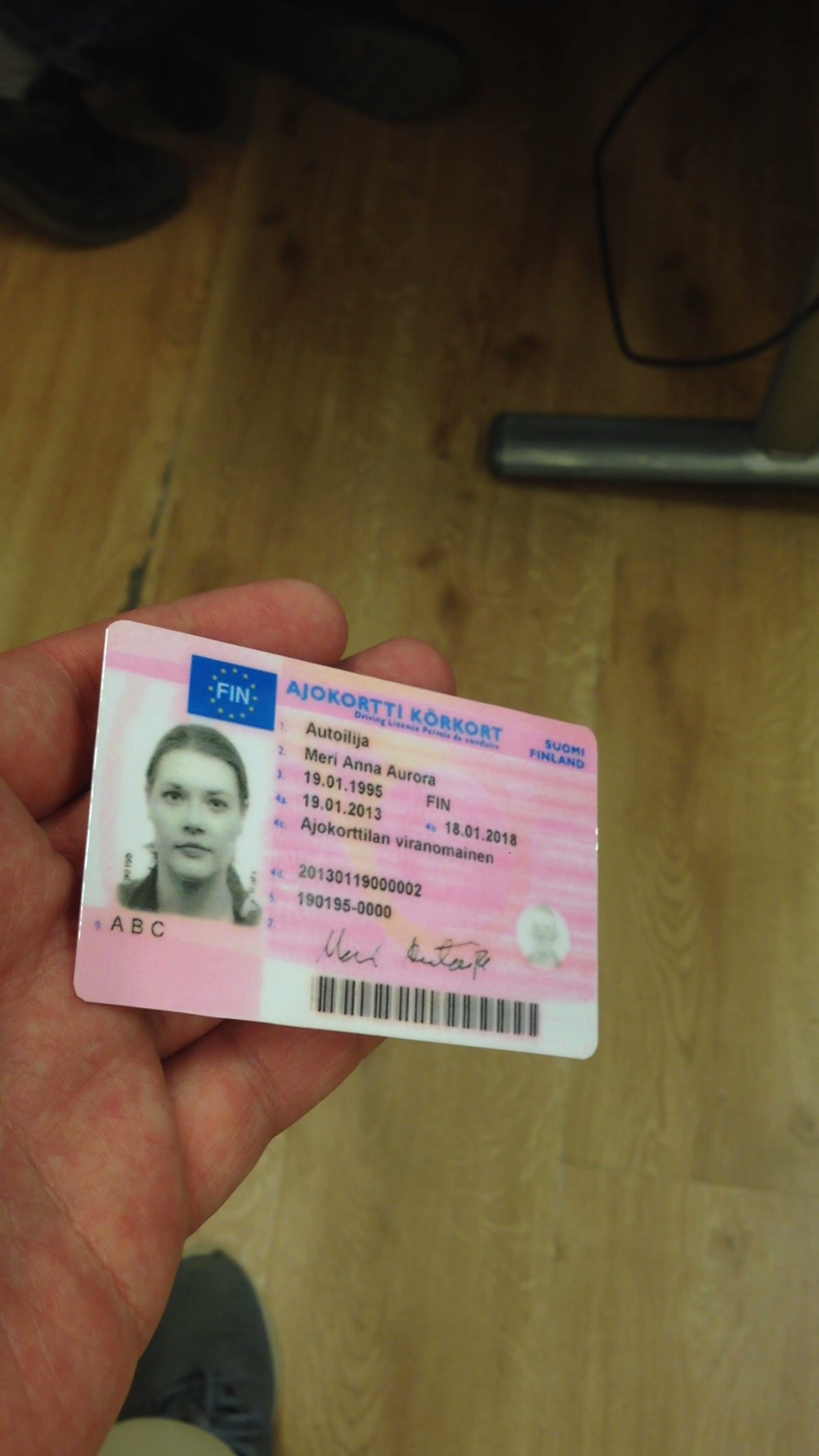}}
\qquad
\subfloat{\includegraphics[width=0.15\textwidth]{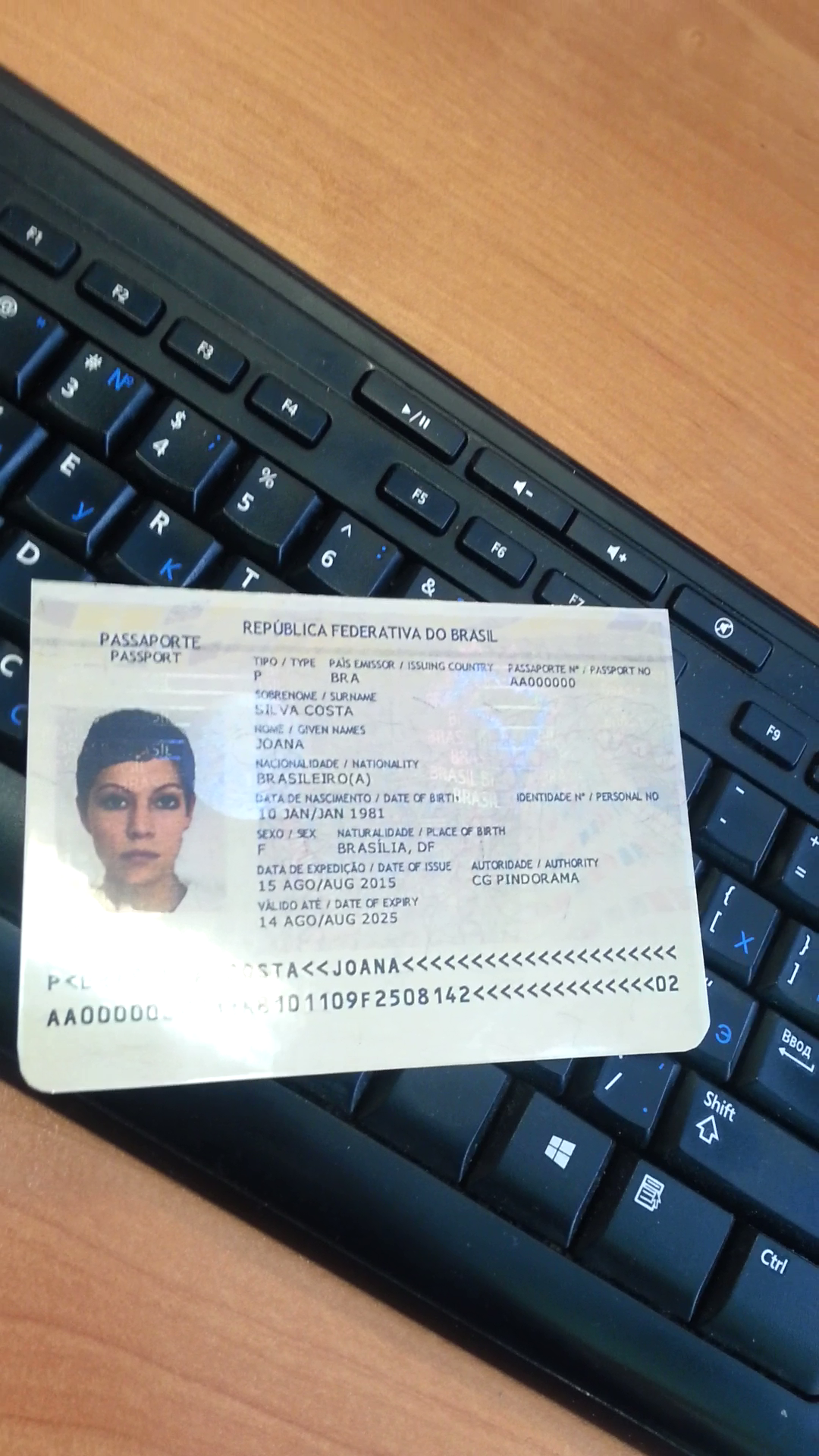}}
\caption{Examples of images from MIDV-500 dataset.}
\label{midv500-samples}
\end{figure}

For our neural network training and evaluation, we use two different open datasets. The first one is the dataset of documents MIDV-500 \cite{midv-500}. This dataset consists of images with different documents captured with mobile devices and which therefore have projective distortion (see examples in Fig. \ref{midv500-samples}). The documents are typically made of plastic or hard paper and therefore planar. In other words, this dataset seems perfect for our task. We use the first 30 types of documents for training and the last 20 types of documents for testing. Also, all the document images with more than one document quadrangle corner outside the image were removed. All valid images were homothetic scaled to a constant width of $400$ before applying the NN.

\begin{figure}[bt]
\centering
\subfloat{\includegraphics[width=0.15\textwidth]{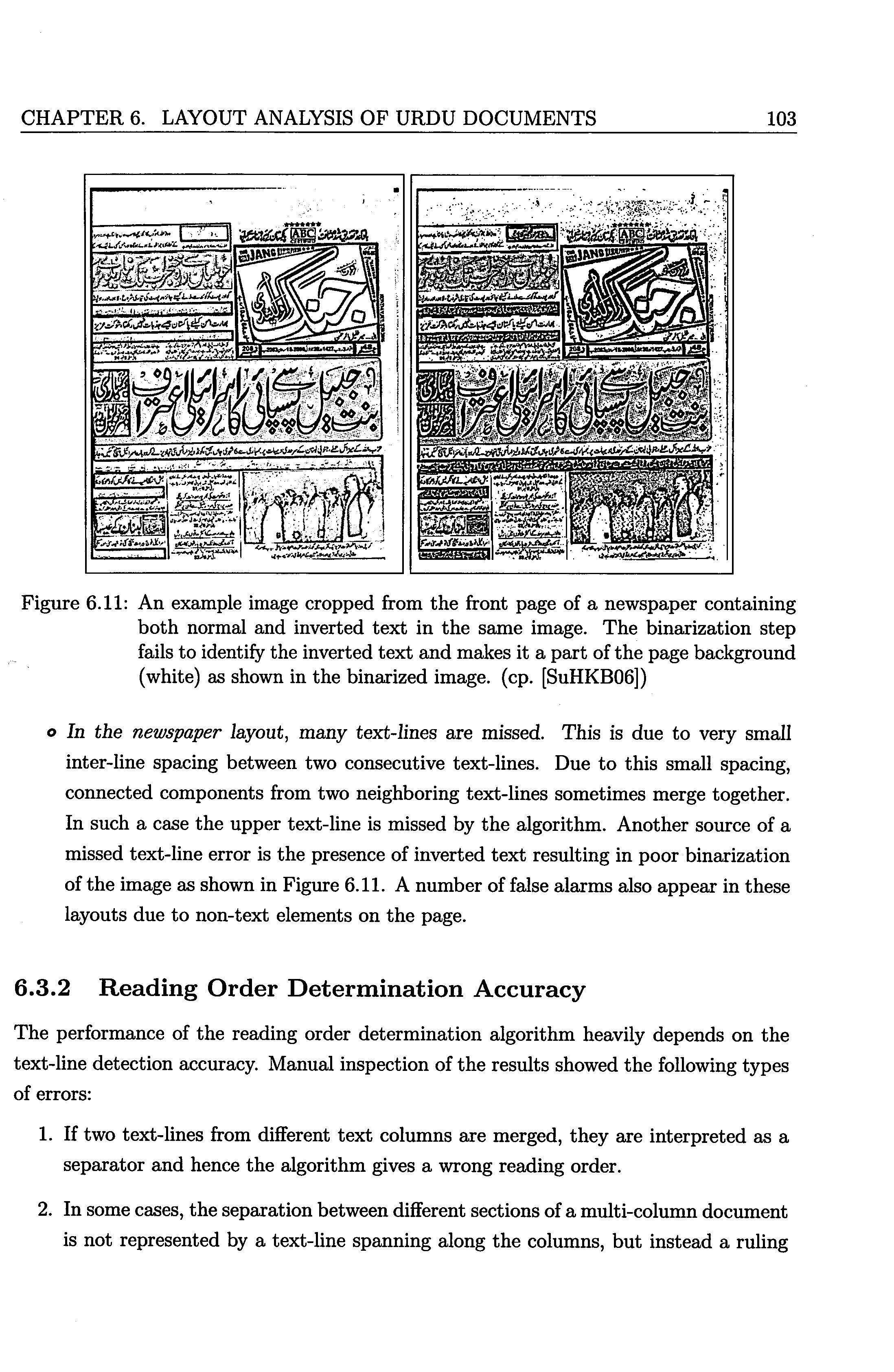}}
\qquad
\subfloat{\includegraphics[width=0.15\textwidth]{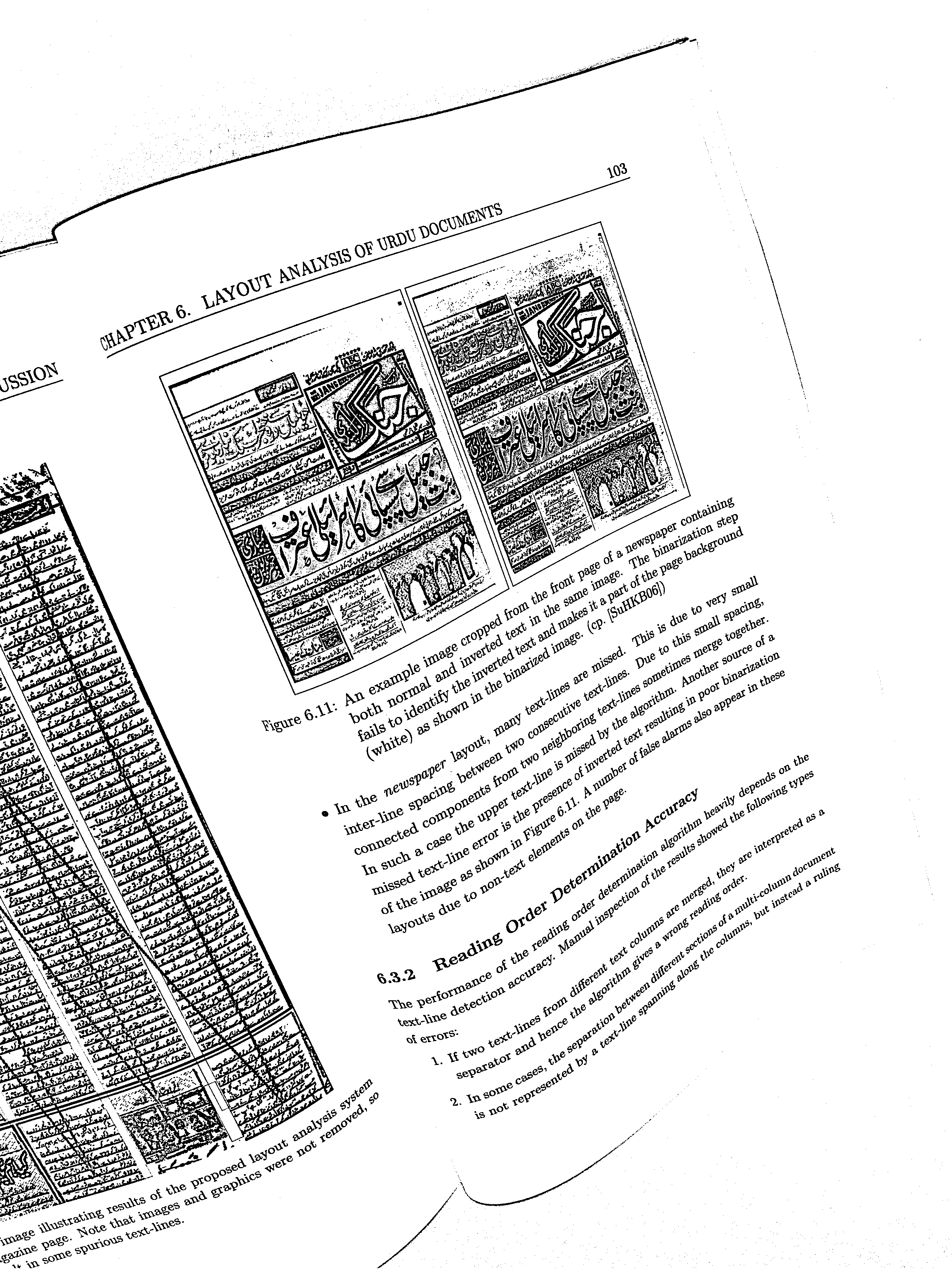}}
\caption{Examples of images from ICDAR 2011 dataset.}
\label{icdar11-samples}
\end{figure}

To evaluate a baseline we use the second dataset from ICDAR 2011 dewarping contest \cite{icdar-2011-data}. These images were scaled to a width of $500$ before the NN usage. Even though the method deals only with projective distortions, its ability to tolerate other distortions is required to use it in a real world tasks. This dataset contains distorted images (projective distortion and page curl) of different pages in binary format (see samples in Fig. \ref{icdar11-samples}). Even having been trained on the different dataset, our NN still outperforms state-of-the-art result which proves strong generalization ability of the suggested approach.

\section{Experimental results}
\label{results}


Our experiments consisted of two parts. For the first part we applied our neural network to the testing part of MIDV-500 dataset and evaluated the quality of vanishing point detection. Using two vanishing points we rectify the images and estimate how rectangular the documents become. To do so we compute two distances: $d_1$ and $d_2$ according to equations ($N$ -- number of the images):

\begin{equation}
\label{metric_1}
d_1=\cfrac{1}{4N} \sum_{j=0}^{N-1}\sum_{i=0}^{3}| 90^\circ - \alpha_i^j|
\end{equation}

\begin{equation}
\label{metric_2}
d_2=\cfrac{1}{2N}\sum_{j=0}^{N-1}|\alpha_v^j| + |\alpha_h^j|
\end{equation}

Distance $d_1$ represents the average deviation from $90^\circ$ of the document rectangle  regardless its orientation (\ref{metric_1}). Distance $d_2$ allows us to estimate how good we managed to correct the orientation of the document (\ref{metric_2}). Table \ref{tab:1} presents results for both corrected and original images to underline the impact of the rectification. We also provide results for the training part to show that our neural network does not suffer from overfitting.


\begin{figure}[bt]
\centering
\includegraphics[width=0.45\textwidth]{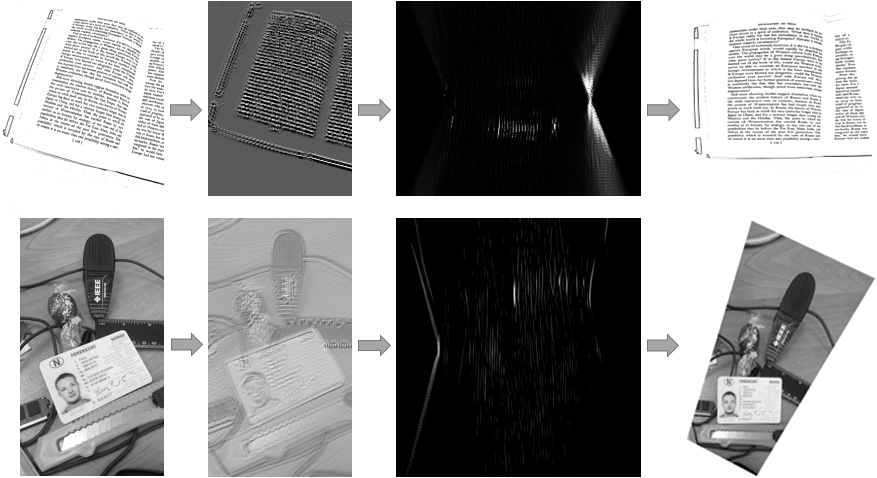}
\caption{Rectification process steps. The first row -- ICDAR 2011 dataset, the second row -- MIDV-500 dataset.}
\label{fht_n_result}
\end{figure}

The second part of the experiment was performed to show both high accuracy in comparison with previously published results and strong generalization ability of our approach. For that purpose we took ICDAR 2011 dewarping contest dataset consisting of 100 binary images and measured quality of recognition by open OCR engine Tesseract (version 3.02)~\cite{tes-act} after the image rectification procedure. We compared our results with \cite{radon-1, radon-2} and it can be shown that even having been trained on completely different images, our method still outperforms the state-of-the-art result (Table \ref{tab:ICDAR_Results}). It was possible because even while the datasets are different, the features (base and cap lines of the text strings, line beginnings and endings, straight elements of the characters) are similar. Another important point is that since these features are not scale invariant, we have to know approximate proportion in the used dataset and scale images respectively.

Fig. \ref{fht_n_result} illustrates the process of image rectification with the proposed neural network. The first column shows the input image, the second column shows the input sample for the first FHT layer (branch for horizontal vanishing point detection), the third one shows the input for the second FHT layer (branch for horizontal vanishing point detection) and the final column shows the image after the rectification process using found vanishing points. We select images from the horizontal branch since illustrations are clearer due to higher linear segments presence. The second and the third columns contain one channel of $12$-channels image which we found to be the most illustrative.  


\begin{table}[bt]
\caption{Results on MIDV-500 dataset}
\label{tab:1}
\begin{tabular}{lllll}
\hline\noalign{\smallskip}
 & Train & Train corrected & Test & Test corrected \\
\noalign{\smallskip}\hline\noalign{\smallskip}
$d_1 (^\circ)$ & 2.56 & \textbf{1.63} & 2.70 & \textbf{1.65} \\
$d_2 (^\circ)$ & 5.76 & \textbf{0.93} & 4.77 & \textbf{0.91} \\
\noalign{\smallskip}\hline
\end{tabular}
\end{table}


\begin{table}[bt]
\caption{Results on ICDAR 2011 dewarping contest dataset}
\label{tab:ICDAR_Results}
\begin{tabular}{lllll}
\hline\noalign{\smallskip}
 & Distorted & By \cite{radon-1} & By \cite{radon-2} & By our method \\
\noalign{\smallskip}\hline\noalign{\smallskip}
Recognized $(\%)$ & 31.3 & 49.6 & 50.1 & \textbf{59.7} \\
\noalign{\smallskip}\hline
\end{tabular}
\end{table}

\section{Conclusion}
\label{conc}

In this paper, we introduce new HoughNet architecture based on usage of the FHT layer which allows convolutional filters to use features from linear areas across the image instead of local areas. Results show very good quality in the task of vanishing points detection in the document images. Our method outperforms the state-of-the-art result on the ICDAR~2011 dewarping contest dataset while used neural network was trained using MIDV-500 dataset. This demonstrates its strong generalization ability. This inspires us to develop this topic further to build a solution for different cases of the task. The suggested approach also show robustness to the document image origin and to complex background. 

For future work, we are planning to develop the idea and expand our solution to all kinds of the vanishing points. Also, we are planning to deal with convergence rate along with accuracy. The other thing to do is to merge neural network branches into one to reduce the amount of trainable parameters.

\section*{Acknowledgment}

This work is partially supported by Russian Foundation for Basic Research (project 18-29-26027 and 17-29-03161).

\IEEEtriggercmd{\enlargethispage{-11in}}
\bibliographystyle{IEEEtran}
\bibliography{bibtex/bib/mybibfile}

\end{document}